\documentclass{article}

\usepackage{arxiv}

\usepackage[utf8]{inputenc} 
\usepackage[T1]{fontenc}    
\usepackage{hyperref}       
\usepackage{url}            
\usepackage{booktabs}       
\usepackage{amsfonts}       
\usepackage{nicefrac}       
\usepackage{microtype}      
\usepackage{lipsum}		
\usepackage{graphicx}
\usepackage{natbib}
\usepackage{doi}
\usepackage{amsmath}

\title{Supervised Stochastic Neighbor Embedding \\Using Contrastive Learning}

\author{ \href{https://orcid.org/0000-0003-3241-3478}{\includegraphics[scale=0.06]{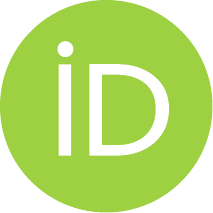}\hspace{1mm}Yi Zhang} \\
	Department of Computer Science \\
	University of Geneva \\
	Geneva, Switzerland \\
	\texttt{yi.zhang.2@etu.unige.ch} \\
}

\renewcommand{\shorttitle}{}

\hypersetup{
pdftitle={},
pdfsubject={ML},
pdfauthor={Yi Zhang, Stéphane Marchand-Maillet},
}

\begin{document}
\maketitle

\begin{abstract}
Stochastic neighbor embedding (SNE) methods $t$-SNE, UMAP are two most popular dimensionality reduction methods for data visualization. Contrastive learning, especially self-supervised contrastive learning (SSCL), has showed great success in embedding features from unlabeled data. The conceptual connection between SNE and SSCL has been exploited. In this work, within the scope of preserving neighboring information of a dataset, we extend the self-supervised contrastive approach to the fully-supervised setting, allowing us to effectively leverage label information. Clusters of samples belonging to the same class are pulled together in low-dimensional embedding space, while simultaneously pushing apart clusters of samples from different classes.
\end{abstract}


\section{Introduction}
Dimensionality reduction (DR) methods map high-dimensional data to a low-dimensional embedding, which enables data visualization. DR methods for visualization have played a critical role to gain insights into high-dimensional data, and the toolkit of DR methods has been rapidly growing in recent years (\cite{mcinnes2020umap}, \cite{sainburg2021parametric}, \cite{amid2022trimap}, \cite{JMLR:v22:20-1061}). Only equiped with a comprehensive understanding of these DR methods, can make informed decision based on data visualization from them, can substantially improve upon them.

The state of the art for unsupervised DR relies on the stochastic neighbor embedding (SNE) framework (\cite{NIPS2002_6150ccc6}), where $t$-SNE (\cite{JMLR:v9:vandermaaten08a}, \cite{JMLR:v15:vandermaaten14a}), UMAP (\cite{mcinnes2020umap}, \cite{sainburg2021parametric}) are two most popular example methods with impressive visualization performance on real-word data. Understanding how these SNE methods work thus has drawn a maasive attention (\cite{JMLR:v23:21-0055}, \cite{JMLR:v22:20-1061}, \cite{damrich2023from}). \cite{JMLR:v22:20-1061} exploited a unified insight into the loss functions of $t$-SNE, UMAP, TriMap, and PaCMAP using graph. \cite{damrich2023from} furthermore generalized negative sampling in  the graph construction, and uncovered the conceptual connection between NE with self-supervised contrastive learning.

In this work, built on the discovered connection (\cite{damrich2023from}), we propose a unified PyTorch framework\footnote{Our code is available: \url{https://github.com/imyizhang/manifold-learn}}, that breaks down the components of $t$-SNE, UMAP, TriMap, and PacMap, and allows to reimplement these SNE methods with self-supervised contrastive setup. Furthermore, given the success of self-supervised contrastive learning, or more generally, contrastive learning (\cite{oord2019representation}, \cite{chen2020simple}, \cite{chen2020big}, \cite{NEURIPS2020_d89a66c7}), we propose a supervised extension to the contrastive SNE methods by leveraging label information, which provides a unified loss function that can be used for either unsupervised or supervised learning.

Our main contributions are summarized as below:
\begin{itemize}
    \item{a unified PyTorch framework for (un)supervised (non-)parametric contrastive NE methods}
    \item a generalized unified loss function
    \item analytic results...
\end{itemize}

\section{Related Work}
Given a dataset of $N$ samples, $\boldsymbol{x}_1, \dots, \boldsymbol{x}_N \in \mathbb{R}^{D}$ in high-dimensional space, dimensionality reduction aims to find low-dimensional embeddings $\boldsymbol{z}_1, \dots, \boldsymbol{z}_N \in \mathbb{R}^{d}$ of input samples, with $D \gg d$  and usually $d=2$ for data visualization.

\subsection{Stochastic Neighbor Embedding}
SNE (\cite{NIPS2002_6150ccc6}) is a powerful representation learning framework that encodes the neighborhood structute and informs the low-dimensional embedding. 

SNE methods usually preserve neighboring information by extracting high-dimensional similarity distribution $P$ over pairs $i j$ of input samples $i$, so-called anchors, and their corresponding nearest neighbors $j$, so-called positives, where $0< i, j \le N$, and then minimizing the lost function between the high-dimensional similarity distribution $P$ and low-dimensional similarity distribution $Q_ \theta$. In a non-parametric setting, the embeddings $\boldsymbol{z}_1, \dots, \boldsymbol{z}_N \in \mathbb{R}^{d}$ become the learnable parameters $\theta$, and $Q_ \theta$ thus become a model. In a paramentric setting, a model is designed to learn a function $f_\theta(\cdot): \mathcal{X} \subset \mathbb{R}^D \rightarrow \mathbb{R}^d$ that encodes $\boldsymbol{x}_i$ into an embedding $\boldsymbol{z}_i$.

The estimation of high-dimensional similarity distribution $P$ differs among the SNE methods. $t$-SNE transforms the Euclidean distances $\operatorname{dist}(\boldsymbol{z}_i, \boldsymbol{z}_j)$ or $d_{i j}$ to the similarities $\operatorname{sim}(\boldsymbol{x}_i, \boldsymbol{x}_j)$ or $p_{i j}$ with a Gaussian kernel, while UMAP transforms with a Laplacian kernel. However, recent work (\cite{NEURIPS2021_2de5d166}, \cite{JMLR:v23:21-0055}) showed that $t$-SNE and UMAP lead barely the same results when using the binary symmetric nearest-neighbor graph in high-dimensional space. We thus use the normalized binary similarities $p_{i j}$ for all SNE methods by default in this work

\begin{equation}
    \operatorname{sim}(\boldsymbol{x}_i, \boldsymbol{x}_j) = p_{i j} = \frac{\mathbb{1}_{i j \in \mathcal{P}}(i j)}{|\mathcal{P}|}
\end{equation}

where $\mathcal{P}$ is the set contains all positive pairs, i.e., $\boldsymbol{x}_i$ is one of the nearest neighbor of $\boldsymbol{x}_j$ or vice versa, $|\mathcal{P}|$ is its cardinality, and $\mathbb{1}$ is the indicator function.

\cite{JMLR:v23:21-0055} also showed that the affects of the different choices of low-dimensional similarity distribution $Q_ \theta$ between $t$-SNE and UMAP are negligible. Therefore, here we transforms the distances $\operatorname{dist}(\boldsymbol{z}_i, \boldsymbol{z}_j)$ or $d_{i j}$ into low-dimensional similarities $\operatorname{sim}(\boldsymbol{z}_i, \boldsymbol{z}_j)$ or $q_{\theta, i j}$ using a Cauchy kernel $\phi(d_{i j})=1 / (d_{i j}^{2} + 1)$ or $\phi_{i j}$ for all SNE methods by default

\begin{equation}
    \operatorname{sim}(\boldsymbol{z}_i, \boldsymbol{z}_j) = q_{\theta, i j} = \phi_{i j}
\end{equation}

$\boldsymbol{t}$\textbf{-SNE}'s lost function measures the Kullback-Leibler divergence of high-dimensional similarity distribution $P$ from low-dimensional similarity distribution $Q_\theta$, which actually requires the normalized low-dimension similarities $q_{\theta, i j} = \phi_{i j} / Z$, where $Z = \sum_{k \neq l} \phi_{k l}$, the partition function. As the entropy of $P$ does not depend on $\theta$, the lost function is equivalent to the expected negative log-likelihood of low-dimensional similarity distribution

\begin{equation}
\begin{split}
    \mathcal{L}_\theta^{t\text{-SNE}} &= -\mathbb{E}_{i j \sim p} \log q_{\theta, i j} \\
    &= -\sum_{i j \in \mathcal{P}} \log \phi_{i j} + \log (\sum_{k l \in \mathcal{P}}\phi_{k l})
\end{split}
\end{equation}

where $p$ is the data distribution.

\textbf{UMAP}'s effective loss function is derived by \cite{NEURIPS2021_2de5d166}

\begin{equation}
\begin{split}
    \mathcal{L}_\theta^{\text{UMAP}} &= -\mathbb{E}_{i j \sim p} \log q_{\theta, i j} -m \mathbb{E}_{i j \sim \xi} \log \left(1-q_{\theta, i j}\right) \\ 
    &= -\sum_{i j \in \mathcal{P}} \log \phi_{i j} -\sum_{i j \in \mathcal{N}} \log \left(1-\phi_{i j}\right) \\
    &= -\sum_{i j \in \mathcal{P}} \log \frac{\tilde{\phi}_{i j}}{\tilde{\phi}_{i j} + 1} -\sum_{i j \in \mathcal{N}} \log \left(1-\frac{\tilde{\phi}_{i j}}{\tilde{\phi}_{i j} + 1}\right)
\end{split}
\end{equation}

where $\xi$ is the approximately uniform noise distribution, and $\mathcal{N}$ is the set of negative pairs, i.e., $\boldsymbol{x}_i$ is approximately uniform sampled from $N$ samples for $\boldsymbol{x}_j$ with $i \neq j$ or vice versa.

\textbf{TriMap}

\begin{equation}
\begin{split}
    \mathcal{L}_\theta^{\text{TriMap}} &= -\mathbb{E}_{i j \sim p} \log q_{\theta, i j} -m \mathbb{E}_{i j \sim \xi} \log \left(1-q_{\theta, i j}\right) \\
    &= -\sum_{i j \in \mathcal{P}, i k \in \mathcal{N}}  \frac{\phi_{i j}}{\phi_{i j} + \phi_{i k}} - w_{U}(t) \sum_{i j \in \mathcal{U}, i k \in \mathcal{U}}  \frac{\phi_{i j}}{\phi_{i j} + \phi_{i k}}
\end{split}
\end{equation}

\textbf{PaCMAP} is another sampling-based SNE method called 

\begin{equation}
\begin{split}
    \mathcal{L}_\theta^{\text{PaCMAP}} &= -\mathbb{E}_{i j \sim p} \log q_{\theta, i j} -m \mathbb{E}_{i j \sim \xi} \log \left(1-q_{\theta, i j}\right) \\
    &= - w_P \sum_{i j \in \mathcal{P}} \frac{\phi_{i j}}{\phi_{i j} + 1}  - w_U(t) \sum_{i j \in \mathcal{U}}  \frac{\phi_{i j}}{\phi_{i j} + 1}  - \sum_{i j \in \mathcal{N}} \left(1- \frac{\phi_{i j}}{\phi_{i j} + 1} \right)
\end{split}
\end{equation}

\subsection{Self-supervised Contrastive Learning}

\textbf{NCE}

\begin{equation}
\begin{split}
    \mathcal{L}_\theta^{\text{NCE}} &= -\mathbb{E}_{i j \sim p} \log q_{\theta, i j} -m \mathbb{E}_{i j \sim \xi} \log \left(1-q_{\theta, i j}\right) \\ 
    &= -\sum_{i j \in \mathcal{P}} \log \frac{\phi_{i j}}{\phi_{i j} + 1} -\sum_{i j \in \mathcal{N}} \log \left(1-\frac{\phi_{i j}}{\phi_{i j} + 1}\right)
\end{split}
\end{equation}

\textbf{InfoNCE}

\begin{equation}
\begin{split}
    \mathcal{L}_\theta^{\text{InfoNCE}} &= -\mathbb{E}_{i j \sim p} \log q_{\theta, i j} -m \mathbb{E}_{i j \sim \xi} \log \left(1-q_{\theta, i j}\right) \\
    &= - \sum_{i j \in \mathcal{P}, i k \in \mathcal{N}}  \log \frac{\operatorname{sim}(\boldsymbol{z}_i, \boldsymbol{z}_j)}{\sum_{i k \in \mathcal{N}} \operatorname{sim}(\boldsymbol{z}_i, \boldsymbol{z}_k)} \\
    &= - \sum_{i j \in \mathcal{P}, i k \in \mathcal{N}}  \log \frac{\phi_{i j}}{\phi_{i j} + \phi_{i k}}
\end{split}
\end{equation}

\textbf{Self-Supervised Contrastive Learning}

\begin{equation}
\begin{split}
    \mathcal{L}_\theta^{\text{Contrastive}} &= -\mathbb{E}_{i j \sim p} \log q_{\theta, i j} -m \mathbb{E}_{i j \sim \xi} \log \left(1-q_{\theta, i j}\right) \\
    &= - \frac{1}{|\mathcal{B}|} \sum_{i \in \mathcal{B}}  \log \frac{\exp \left( \operatorname{sim}(\boldsymbol{z}_i, \boldsymbol{z}_j) / \tau \right)}{\sum_{i k \in \mathcal{N}} \exp \left( \operatorname{sim}(\boldsymbol{z}_i, \boldsymbol{z}_k) / \tau 
    \right)} \\
\end{split}
\end{equation}

\textbf{Soft Nearest Neighbors}

\begin{equation}
\begin{split}
    \mathcal{L}_\theta^{\text{Soft Nearest Neighbors}} &= -\mathbb{E}_{i j \sim p} \log q_{\theta, i j} -m \mathbb{E}_{i j \sim \xi} \log \left(1-q_{\theta, i j}\right) \\
    &= - \frac{1}{|\mathcal{B}|} \sum_{i \in \mathcal{B}}  \log \sum_{i j \in \mathcal{P}} \frac{\exp \left( \operatorname{sim}(\boldsymbol{z}_i, \boldsymbol{z}_j) / \tau \right)}{\sum_{i k \in \mathcal{N}} \exp \left( \operatorname{sim}(\boldsymbol{z}_i, \boldsymbol{z}_k) / \tau 
    \right)} \\
\end{split}
\end{equation}

\subsection{Supervised Contrastive Learning}

\textbf{Supervised Contrastive Learning}

\begin{equation}
\begin{split}
    \mathcal{L}_\theta^{\text{Contrastive}} &= -\mathbb{E}_{i j \sim p} \log q_{\theta, i j} -m \mathbb{E}_{i j \sim \xi} \log \left(1-q_{\theta, i j}\right) \\
    &= - \frac{1}{|\mathcal{B}|} \sum_{i \in \mathcal{B}} \left( \frac{1}{|\tilde{\mathcal{P}}|} \sum_{i j \in \tilde{\mathcal{P}}} \log \frac{\exp \left( \operatorname{sim}(\boldsymbol{z}_i, \boldsymbol{z}_j) / \tau \right)}{\sum_{i k \in \mathcal{N}} \exp \left( \operatorname{sim}(\boldsymbol{z}_i, \boldsymbol{z}_k) / \tau 
    \right)} \right)\\
\end{split}
\end{equation}

\textbf{Soft nearest neighbor}

\begin{equation}
\begin{split}
    \mathcal{L}_\theta^{\text{Soft Nearest Neighbors}} &= -\mathbb{E}_{i j \sim p} \log q_{\theta, i j} -m \mathbb{E}_{i j \sim \xi} \log \left(1-q_{\theta, i j}\right) \\
    &= - \frac{1}{|\mathcal{B}|} \sum_{i \in \mathcal{B}}  \log \left( \frac{1}{|\tilde{\mathcal{P}}|} \sum_{i j \in \tilde{\mathcal{P}}} \frac{\exp \left( \operatorname{sim}(\boldsymbol{z}_i, \boldsymbol{z}_j) / \tau \right)}{\sum_{i k \in \mathcal{N}} \exp \left( \operatorname{sim}(\boldsymbol{z}_i, \boldsymbol{z}_k) / \tau 
    \right)} \right)\\
\end{split}
\end{equation}

\section{Method}

\begin{equation}
\begin{split}
    \mathcal{L}_\theta^{t\text{-SCNE}} &= -\mathbb{E}_{i j \sim p} \log q_{\theta, i j} -m \mathbb{E}_{i j \sim \xi} \log \left(1-q_{\theta, i j}\right) \\
    &= \frac{1}{|\mathcal{B}|} \sum_{i \in \mathcal{B}} \left( -\frac{1}{|\tilde{\mathcal{P}}|} \sum_{i j \in \tilde{\mathcal{P}}} \frac{\phi_{i j}}{\sum_{i k \in \mathcal{N}} \phi_{i k}} - w_{U}(t) \frac{\phi_{i j}}{\sum_{i k \in \mathcal{U}} \phi_{i k}} \right)\\
\end{split}
\end{equation}

\section{Experiments}

\section{Discussion}

\bibliographystyle{unsrtnat}
\bibliography{references}  






\end{document}